\documentclass[11pt,a4paper]{article}
\usepackage[hyperref]{acl2021}
\usepackage{times}
\usepackage{latexsym}
\usepackage{booktabs}
\usepackage{caption}
\usepackage{graphicx, subfig}
\usepackage{multirow}
\usepackage{float}
\usepackage{amsmath} 

\usepackage{microtype}

\aclfinalcopy 
\def\aclpaperid{1895} 

\title{Bilingual Mutual Information Based Adaptive Training \\ for Neural Machine Translation}

\author{
  Yangyifan Xu\textsuperscript{1}\thanks{\ \ This work was done when Yangyifan Xu was interning at Pattern Recognition Center, WeChat AI, Tencent Inc, China} \ ,
  Yijin Liu\textsuperscript{1,2}, 
  Fandong Meng\textsuperscript{2}, 
  Jiajun Zhang\textsuperscript{3,4},
  Jinan Xu\textsuperscript{1}\thanks{ \ \ Jinan Xu is the corresponding author of the paper.}  
  and Jie Zhou\textsuperscript{2} \\
  \textsuperscript{1}Beijing Jiaotong University, China \\
  \textsuperscript{2}Pattern Recognition Center, WeChat AI, Tencent Inc, China \\
  \textsuperscript{3}National Laboratory of Pattern Recognition, Institute of Automation, CAS \\
  \textsuperscript{4}School of Artificial Intelligence, University of Chinese Academy of Sciences \\
  \texttt{xuyangyifan2021@ia.ac.cn}, \texttt{adaxry@gmail.com} \\
  \texttt{\{fandongmeng, withtomzhou\}@tencent.com} \\
  \texttt{jjzhang@nlpr.ia.ac.cn}, \texttt{jaxu@bjtu.edu.cn} \\
}

\date{}

\begin{document}
\maketitle
\begin{abstract}
Recently, token-level adaptive training has achieved promising improvement in machine translation, where the cross-entropy loss function is adjusted by assigning different training weights to different tokens, in order to alleviate the token imbalance problem.
However, previous approaches only use static word frequency information in the target language without considering the source language, which is insufficient for bilingual tasks like machine translation.
In this paper, we propose a novel bilingual mutual information (BMI) based adaptive objective, which measures 
the learning difficulty for each target token from the perspective of bilingualism, and assigns an adaptive weight accordingly to improve token-level adaptive training.
This method assigns larger training weights to tokens with higher BMI, so that easy tokens are updated with coarse granularity while difficult tokens are updated with fine granularity. 
Experimental results on WMT14 English-to-German and WMT19 Chinese-to-English demonstrate the superiority of our approach compared with the 
Transformer baseline and previous token-level adaptive training approaches.
Further analyses confirm that our method can improve the lexical diversity.
\end{abstract}

\section{Introduction}
Neural machine translation (NMT) \citep{cho2014learning,NIPS2014seq2seq, BahdanauCB14,vaswani2017attention,chen2018best,meng2019dtmt,zhang2019bridging,yanetal2020multi,liu2021faster} has achieved remarkable success.
As a data-driven model, 
the performance of NMT depends on training corpus.
Balanced training data is a crucial factor in building a superior model.
However, natural languages conform to the Zipf's law \citep{zipf1949human}, the frequencies of words exhibit the long tail characteristics, which brings an imbalance in the distribution of words in training corpora. 
Some studies \citep{jiang2019improving, gu2020token} assign different training weights to target tokens according to their frequencies. 
These approaches alleviate the token imbalance problem and indicate that tokens should be treated differently during training.

However, there are two issues in existing approaches. 
First, these approaches believe that low-frequency words are not sufficiently trained and thus amplify the weight of them. Nevertheless, low-frequency tokens are not always difficult as the model competence increases \citep{wan2020self}.
Second, previous studies only use monolingual word frequency information in the target language without considering the source language, which is insufficient for bilingual tasks, e.g., machine translation. 
The mapping between bilingualism is a more appropriate indicator.
As shown in Table \ref{fig:example}, word frequency of \textit{pleasing} and \textit{bearings} are both 847. Corresponding to Chinese, \textit{pleasing} has multiple mappings, while \textit{bearings} is relatively single.
The more multivariate the mapping is, the less confidence in predicting the target word given the source context.
\citet{he2019towards} also confirm this view that words with multiple mappings contribute more to the BLEU score.

\begin{table}[!t]
\setlength{\tabcolsep}{3mm}{
\begin{tabular}{c|l}
\hline
pleasing (847) & \begin{tabular}[c]{@{}l@{}}g\={a}ox\`{i}ng (81); y\'{u}ku\`{a}i (74);\\ xǐyu\`{e} (63); qǔyu\`{e }(49) ...\end{tabular} \\ \hline
bearings (847) & zh\'{o}uch\'{e}ng (671) ...                                \\ \hline
\end{tabular}}
\caption{\label{fig:example} An example from the WMT19 Chinese-English training set. The Chinese words are presented in pinyin style and the word frequency is shown in brackets. The two words have the same word frequency, while the mapping of \textit{bearings} is more stable than that of \textit{pleasing}.}
\end{table}

To tackle the above issues, we propose bilingual mutual information (BMI), which has two characteristics:
1) BMI measures the learning difficulty for each target token by considering the strength of association between it and the source sentence; 2) for each target token, BMI can dynamically adjust according to the context.
BMI-based adaptive training can dynamically adjust the learning granularity on tokens. Easy tokens are updated with coarse granularity while difficult tokens are updated with fine granularity.

We evaluate our approach on both WMT14 English-to-German and WMT19 Chinese-to-English translation tasks. 
Experimental results on two benchmarks demonstrate the superiority of our approach compared with the 
Transformer baseline and previous token-level adaptive training approaches.
Further analyses confirm that our method can improve the lexical diversity. The main contributions\footnote{Reproducible code: https://github.com/xydaytoy/BMI-NMT} of this paper can be summarized as follows:

\begin{itemize}
\item We propose a 
training objective based on bilingual mutual information (BMI),
which can reflects the learning difficulty for each target token from the perspective of bilingualism, and assigns an adaptive weight accordingly to guide the adaptive training of machine translation.
\item Experimental results show that our method can improve not only the machine translation quality, but also the lexical diversity.
\end{itemize}

\section{Background}
\subsection{Neural Machine Translation}
A 
NMT system is a neural network that translates a source sentence $\mathbf{x}$ with $n$ words to a target sentence $\mathbf{y}$ with $m$ words.
During the training process, NMT models are optimized by minimizing cross entropy:
\begin{equation}
\mathcal{L} = -\frac{1}{m}\sum_{j=1}^{m}{\rm log}p(y_{j}|\mathbf{y}_{<j},\mathbf{x}), \label{con:CEloss}
\end{equation}
where 
$y_j$ 
is the ground-truth token at the $j$-th position and $\mathbf{y}_{<j}$ is the translation history known before predicting token $y_j$.

\subsection{Token-level Adaptive Training Objective}

Following \citep{gu2020token}, the token-level adaptive training objective is
\begin{equation}
\mathcal{L} = -\frac{1}{m}\sum_{j=1}^{m}w_j\cdot {\rm log}p(y_{j}|\mathbf{y}_{<j},\mathbf{x}),
\label{con:weightedloss}
\end{equation}
where $w_j$ is the weight assigned to the target token $y_{j}$ .
\citet{gu2020token} used monolingual word frequency information in the target language to calculate the $w_j$. 
The weight does not contain the information of the source language, and cannot be dynamically adjusted with the context.

\begin{figure}[!t]
\scalebox{0.95}{
\centering
\includegraphics[width=8cm]{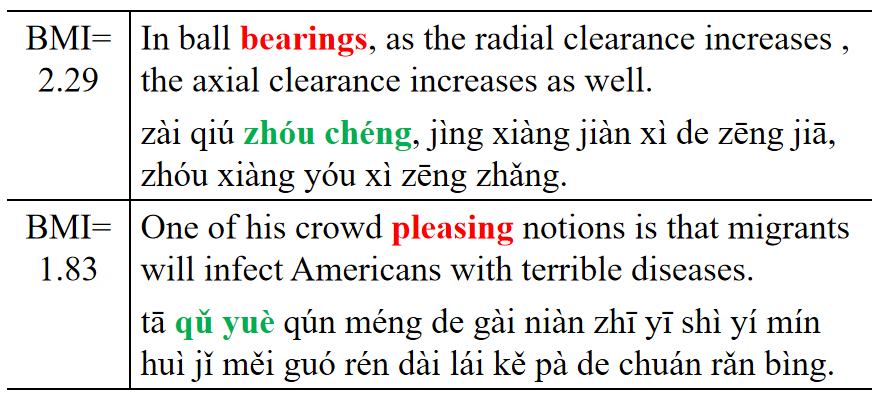}}
\caption{An example from WMT19 Chinese-to-English training set. Words with Red and Bold fonts have the same word frequency while different BMI.}
\label{fig:running}
\end{figure}

\section{BMI-based Adaptive Training}
In this section, we start with the definition of the bilingual mutual information (BMI). 
Then we analyze the relationship between BMI and translation difficulty. 
Based on this, we introduce our BMI-based token-level adaptive training objective.

\subsection{Definition of BMI}
Mutual information measures the strength of association between two random variables by comparing the number of their individual and joint occurrences.
We develop BMI, which is calculated by summarizing the mutual information of the target token and each token in the source sentence, to measure the 
learning difficulty of the model.
Token pairs with high BMI are considered easy, since they have high co-occurrence relative to the frequency. 
Given the source sentence $\mathbf{x}$ and target token $y_j$, we define the bilingual mutual information as\footnote{To ensure comparability of two probability distribution, the tokens that appear multiple times in a sentence and the token pairs that appear multiple times in a sentence pair are not counted repeatedly.}:
\begin{equation}
{\rm BMI}(\mathbf{x},y_j)=\sum_{i=1}^{n} {\rm log}\frac{f(x_i,y_j)}{f(x_i) \cdot f(y_j)/K},
\label{con:BMI}
\end{equation}
where $f(x_i)$ and $f(y_j)$ are total number of sentences in the corpus containing at least one occurrence of $x_i$ and $y_j$, respectively, $f(x_i, y_j)$ represents total number of sentences in the corpus having at least one occurrence of the word pair ($x_i$, $y_j$), 
and $K$ denotes total number of sentences in the corpus.

\begin{figure}[!t]
\scalebox{0.9}{
\centering
\includegraphics[width=8cm]{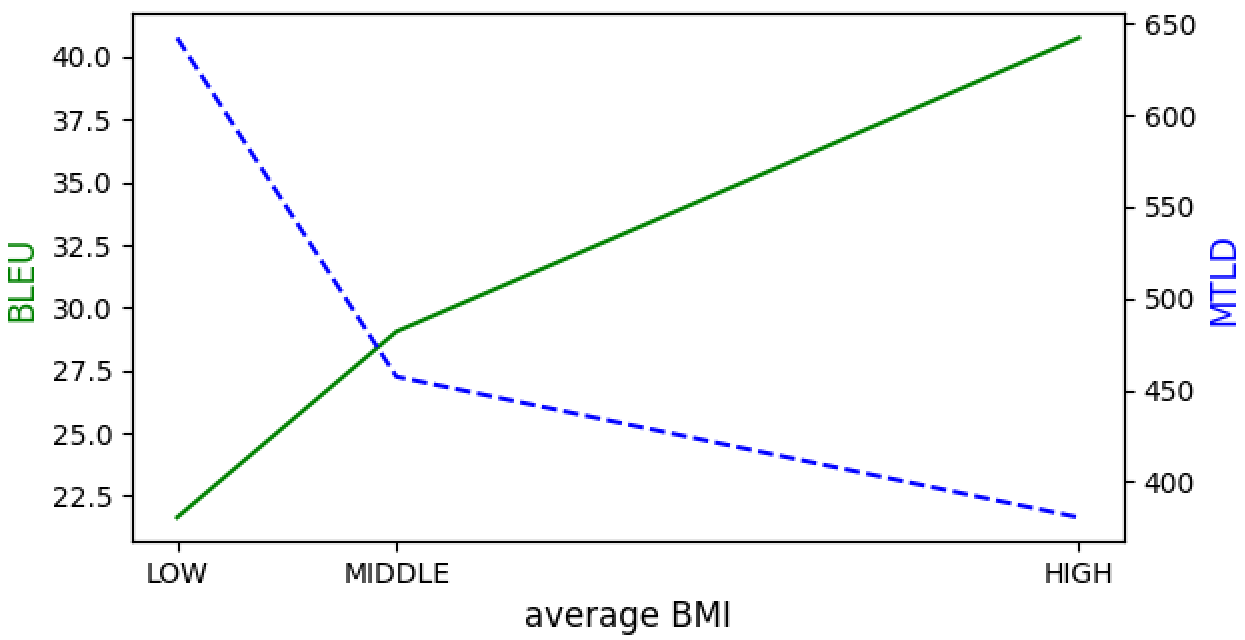} }
\caption{The BLEU (green solid line) and MTLD (blue dotted line) values on the subsets of WMT14 English-German training set,  divided according to the average BMI.
All target sentences of the training set are divided into three subsets according to the average BMI of the tokens in the sentence, which are equal in size and denoted as LOW, MIDDLE, and HIGH, respectively.
BLEU indicates the learning difficulty of the model.
The measure of textual lexical diversity (MTLD) \citep{mccarthy2010mtld} represents the lexical diversity of the data set.
The results show that high BMI means relatively stable mapping, which is easy to be learned by the model and has low lexical diversity.}
\label{fig:bleuandmi}
\end{figure}

\subsection{What BMI Measures?}
We use an example to illustrate our idea.
Figure \ref{fig:running} shows two sentence pairs. Words with Red and Bold fonts have the same word frequency. As shown in Table \ref{fig:example}, \textit{pleasing} has multiple mappings, while the mapping of \textit{bearings} is relatively single. As a result, the appearance of corresponding English word brings different confidence of the appearance of the Chinese word, which can be reflected by BMI.
Further statistical results are shown in Figure \ref{fig:bleuandmi}, high BMI means relatively stable mapping, which is easy to be learned by the model and has low lexical diversity.

\subsection{BMI-based Objective}
We calculate the token-level weight by scaling BMI and adjusting the lower limit as follows:
\begin{equation}
w_j=S\cdot {\rm BMI}(\mathbf{x},y_j)+B.
\label{con:BMIloss}
\end{equation}
The two hyperparameters $S$ (scale) and $B$ (base) influence the magnitude of change and the lower limit, respectively.



In training process, the loss of simple tokens will be amplified, the model updates simple tokens with coarse granularity, because our strategy thinks the model can easily predict these target tokens given the source sentence, and it needs to increase the penalty if the prediction is wrong.
For difficult tokens, the model has a higher tolerance because their translation errors may not be absolute. As a result, the loss is small due to the small weight and the difficult tokens are always updated in a fine-grained way.

\section{Experiments}
\label{sec:Experiments}

We evaluate our method on the 
Transformer \citep{vaswani2017attention} and conduct experiments on two widely-studied NMT tasks, WMT14 English-to-German (En-De) and WMT19 Chinese-to-English (Zh-En).
\subsection{Data Preparation}
\label{ssec:Data Preparation}

\paragraph{EN-DE.}
The training data consists of 4.5M sentence pairs from WMT14. Each word in the corpus has been segmented into subword units using byte pair encoding (BPE) \citep{sennrich2016neural} with 32k merge operations. The vocabulary is shared among source and target languages. We select newstest2013 for validation and report the BLEU scores on newstest2014.

\paragraph{ZH-EN.}
The training data is from WMT19 which consists of 20.5M sentence pairs.
The number of merge operations in byte pair encoding (BPE) is set to 32K for both source and target languages.
We use newstest2018 as our validation set and newstest2019 as our test set, which contain 4k and 2k sentences, respectively.

\subsection{Systems}
\paragraph{Transformer.}
We implement our approach with the open source toolkit THUMT \citep{zhang2017thumt} and strictly follow the setting of Transformer-Base in \cite{vaswani2017attention}.

\paragraph{Exponential \citep{gu2020token}.}
This method adds an additional training weights to low-frequency target tokens:
\begin{equation}
w_j=A\cdot e^{-T\cdot Count(y_j)}+1.
\label{con:EXP}
\end{equation}

\paragraph{Chi-Square \citep{gu2020token}.}
The weighting function of this method is similar to the form of chi-square distribution
\begin{equation}
w_j=A\cdot Count^2(y_j)e^{-T\cdot Count(y_j)}+1.
\label{con:CHI}
\end{equation}

\paragraph{BMI.}
Our system is first trained with normal cross entropy loss (Equation \ref{con:CEloss}) for 100k steps. 
Then the model is further trained with BMI-based adaptive objective (Equation \ref{con:BMIloss}) for 100k steps. The same procedure was used for the competing methods. In order to eliminate the influence of noise, we assign the weight of tokens with BMI lower than 0.4 to zero during the training process. 

\subsection{Hyperparameters}


We introduce two hyperparameters, $B$ and $S$, to adjust the weight distribution based on BMI, as shown in Equation \ref{con:BMIloss}. 
In our experiments, we fixed $B$ to narrow search space [0.7, 1]. We tuned another hyperparameter $S$ on the validation sets. The results are shown in Table \ref{table:hyperparameter_appendix}. 
Finally, we use the best hyperparameters found on the validation set for the final evaluation of the test set. For En-De, $B$ = 0.8 and $S$ = 0.15, for Zh-En, $B$ = 1.0 and $S$ = 0.1.

\begin{table}[!tbp]
\setlength{\tabcolsep}{3mm}{
\begin{tabular}{c|c|c|c|c}
\hline
\textbf{}             & \textbf{$B$}           & \textbf{$S$} & \textbf{EN-DE} & \textbf{ZH-EN} \\ \hline
\multirow{15}{*}{BMI} & \multirow{6}{*}{1.0} & 0.05       & 26.87          & 23.52          \\ \cline{3-5} 
                      &                      & 0.10       & 26.89          & \textbf{23.61} \\ \cline{3-5} 
                      &                      & 0.15       & 26.93          & 23.49          \\ \cline{3-5} 
                      &                      & 0.20       & 26.98          & 23.39          \\ \cline{3-5} 
                      &                      & 0.25       & 26.91          & 23.24          \\ \cline{3-5} 
                      &                      & 0.30       & 26.85          & 23.50          \\ \cline{2-5} 
                      & \multirow{3}{*}{0.9} & 0.15       & 26.93          & 23.31               \\ \cline{3-5} 
                      &                      & 0.20       & 26.88          & 23.31                \\ \cline{3-5} 
                      &                      & 0.25       & 26.96          & 23.41                \\ \cline{2-5} 
                      & \multirow{3}{*}{0.8} & 0.15       & \textbf{27.01}          & 23.40               \\ \cline{3-5} 
                      &                      & 0.20       & 26.81          & 23.25               \\ \cline{3-5} 
                      &                      & 0.25       & 26.93          & 23.50               \\ \cline{2-5} 
                      & \multirow{3}{*}{0.7} & 0.15       & 26.92          & 23.44               \\ \cline{3-5} 
                      &                      & 0.20       & 26.90          & 23.35               \\ \cline{3-5} 
                      &                      & 0.25       & 26.89          & 23.34               \\ \hline
\end{tabular}}
\caption{\label{table:hyperparameter_appendix} Performance of our methods on the validation sets with different hyperparameters $S$ and $B$. }
\end{table}


\begin{table}[!t]
\setlength{\tabcolsep}{1mm}{
\scalebox{0.9}{
\begin{tabular}{lccll}
\cline{1-3}
\multicolumn{1}{l|}{\textbf{System}} & \multicolumn{1}{c|}{\textbf{EN-DE}} & \textbf{ZH-EN} &  &  \\ \cline{1-3}
\multicolumn{3}{c}{\textit{Existing NMT systems}}                                           &  &  \\ \cline{1-3}
\multicolumn{1}{l|}{\citet{vaswani2017attention}}               & \multicolumn{1}{c|}{27.3}           & -              &  &  \\ \cline{1-3}
\multicolumn{1}{l|}{Chi-Square}          & \multicolumn{1}{c|}{27.51}          & -              &  &  \\ \cline{1-3}
\multicolumn{1}{l|}{Exponential}         & \multicolumn{1}{c|}{27.60}           & -              &  &  \\ \cline{1-3}
\multicolumn{3}{c}{\textit{Our NMT systems}}                                                &  &  \\ \cline{1-3}
\multicolumn{1}{l|}{Transformer}        & \multicolumn{1}{c|}{27.97}          & 24.37          &  &  \\ \cline{1-3}
\multicolumn{1}{l|}{ ~~~~~~~~+ Chi-Square}          & \multicolumn{1}{c|}{28.08(\textit{+0.11})}   & 24.62(\textit{+0.25})   &  &  \\ \cline{1-3}
\multicolumn{1}{l|}{~~~~~~~~+ Exponential}         & \multicolumn{1}{c|}{28.17(\textit{+0.20})}    & 24.33(\textit{-0.04})   &  &  \\ \cline{1-3}
\multicolumn{1}{l|}{~~~~~~~~+ BMI}             & \multicolumn{1}{c|}{\textbf{28.53(\textit{+0.56})}*}   & \textbf{25.19(\textit{+0.82})*}   &  &  \\ \cline{1-3}
\end{tabular}} }
\caption{\label{table:mainresult} BLEU scores (\%) on the WMT14 En-De test set and the WMT19 Zh-En test set. Results of our method marked with `*' are statistically significant \cite{koehn2004statistical} by contrast to all other models (p\textless0.01).}
\end{table}

\begin{figure}[!t]
\scalebox{0.95}{
\centering
\includegraphics[width=8cm]{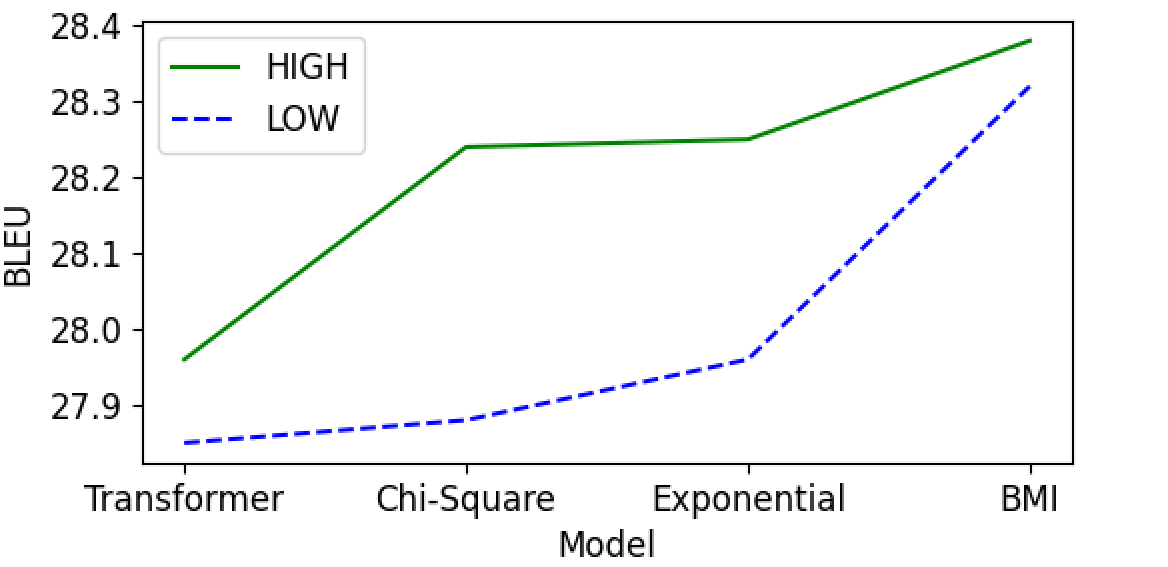}}
\caption{BLEU scores (\%) on different WMT14 En-De test subsets which are grouped by their average BMI. Sentences in the HIGH subset contains more tokens with high BMI. 
}
\label{fig:highandlow}
\end{figure}

\subsection{Main Results}
As shown in Table \ref{table:mainresult}, compared  with \cite{vaswani2017attention}, our Transformer outperforms it by 0.67 BLEU points.
We use a strong baseline system in this work in order to make the evaluation convincing.
Improvement of existing methods \cite{gu2020token} is limited over strong baseline. Exponential objective achieves 28.17 (+0.2) BLEU on En-De and Chi-Square objective achieves 24.62 (+0.25) BLEU on Zh-En.
Our method yields 28.53 (+0.56) and 25.19 (+0.82) BLEU on the En-De task and Zh-En task, respectively. The significant and consistent improvement on the two large-scale dataset demonstrates the effectiveness of our method.

\subsection{Results on Different BMI Intervals}
We score each target sentence of newstest2014 by calculating the average BMI of each token in the sentence, and then divide newstest2014 into two subsets with equal size according to the score, denoted as HIGH and LOW, respectively.
As shown in Figure \ref{fig:highandlow}, compared to Transformer, frequency-based methods outperform on the HIGH subset but have no obvious improvement on the LOW subset.
By contrast, our method can not only bring a stable improvement on the HIGH subset, the improvement is even more obvious on the LOW subset.
Low BMI means relatively rich mapping. We believe that the model should have a higher tolerance for these tokens because their translation errors may not be absolute. For example, the model outputs another token with similar meaning. Therefore, our method improves more on LOW subset.


\subsection{Effects on Lexical Diversity}
\citet{vanmassenhove2019lost} suggest that the vanilla NMT systems exacerbate bias presented in corpus, resulting in lower vocabulary diversity.
We use three measures of lexical diversity, namely, moving-average type-token ratio (MATTR) \citep{covington2010cutting}, the approximation of hypergeometric distribution (HD-D) and the measure of textual lexical diversity (MTLD) \citep{mccarthy2010mtld}. Results in Table \ref{table:diver} show that, on improving the lexical diversity of translation, our method is superior to  existing methods (Chi-Square and Exponential) based on word frequency.

\begin{table}[!t]
\centering
\setlength{\tabcolsep}{2mm}{
\scalebox{0.95}{
\begin{tabular} {l|c|c|c}
\hline \textbf{Models} & \textbf{MATTR} & \textbf{HD-D} & \textbf{MTLD} \\ \hline
Transformer & 89.41 & 94.05 & 230.36 \\
~~~~~~~+ Chi-Square & 89.37 & 94.02 & 230.02 \\
~~~~~~~+ Exponential & 89.41 & 94.08 & 232.98 \\
~~~~~~~+ BMI & \textbf{89.45} & \textbf{94.10} & \textbf{236.43} \\
Reference & 90.92 & 94.88 & 259.98 \\
\hline
\end{tabular}} }
\caption{\label{table:diver} The lexical diversity of WMT14 En-De translations measured by MATTR (\%), HD-D (\%) and MTLD. A larger value means a higher diversity. 
}
\end{table}

\subsection{Contrast with Label Smoothing}
There are similarities between token-level adaptive training and label smoothing, because they both adjust the loss function of the model by token weighting. In particular, for some smoothing methods guided by prior or posterior knowledge of training data \citep{gao2020towards,pereyra2017regularizing}, different tokens are treated differently. But these similarities are not the key points of the two methods, and they are essentially different. The first and very important point is that the motivations of the two methods are different. Label smoothing is a regularization method to avoid overfitting, while our method treats samples of different difficulty differently for adaptive training. Second, the two methods work in different ways. Label smoothing is used when calculating the cross-entropy loss. It emphasizes how to assign the weight of tokens other than the golden one, and indirectly affects the training of the golden token. While our method is used after calculating the cross-entropy loss. It is calculated according to the golden token at each position in the reference, which is more direct. In all experiments, we employed uniform label smoothing of value $\epsilon_{ls}$ = 0.1, the results show that the two methods does not conflict when used together.

\section{Conclusion}
We propose a novel bilingual mutual information based adaptive training objective, which can measure the learning difficulty for each target token from the perspective of bilingualism, and adjust the learning granularity dynamically to improve token-level adaptive training. Experimental results on two translation tasks show that our method can bring a significant improvement in translation quality,  especially  on  sentences that are difficult to learn by the model. Further analyses confirm that our method can also improve the lexical diversity.

\section*{Acknowledgments}
The research work descried in this paper has been supported by the National Key R\&D Program of China (2020AAA0108001), the National Nature Science Foundation of China (No. 61976015, 61976016, 61876198 and  61370130) and the Beijing Academy of Artificial Intelligence (BAAI2019QN0504). The authors would like to thank the anonymous reviewers for their valuable comments and suggestions to improve this paper.

\bibliographystyle{acl_natbib}
\bibliography{anthology,acl2021}

\begin{thebibliography}{22}
\expandafter\ifx\csname natexlab\endcsname\relax\def\natexlab#1{#1}\fi

\bibitem[{Bahdanau et~al.(2015)Bahdanau, Cho, and Bengio}]{BahdanauCB14}
Dzmitry Bahdanau, Kyunghyun Cho, and Yoshua Bengio. 2015.
\newblock \href {http://arxiv.org/abs/1409.0473} {Neural machine translation by
  jointly learning to align and translate}.
\newblock In \emph{3rd International Conference on Learning Representations,
  {ICLR} 2015, San Diego, CA, USA, May 7-9, 2015, Conference Track
  Proceedings}.

\bibitem[{Chen et~al.(2018)Chen, Firat, Bapna, Johnson, Macherey, Foster,
  Jones, Schuster, Shazeer, Parmar et~al.}]{chen2018best}
Mia~Xu Chen, Orhan Firat, Ankur Bapna, Melvin Johnson, Wolfgang Macherey,
  George Foster, Llion Jones, Mike Schuster, Noam Shazeer, Niki Parmar, et~al.
  2018.
\newblock The best of both worlds: Combining recent advances in neural machine
  translation.
\newblock In \emph{Proceedings of the 56th Annual Meeting of the Association
  for Computational Linguistics (Volume 1: Long Papers)}, pages 76--86.

\bibitem[{Cho et~al.(2014)Cho, van Merri{\"e}nboer, Gulcehre, Bahdanau,
  Bougares, Schwenk, and Bengio}]{cho2014learning}
Kyunghyun Cho, Bart van Merri{\"e}nboer, Caglar Gulcehre, Dzmitry Bahdanau,
  Fethi Bougares, Holger Schwenk, and Yoshua Bengio. 2014.
\newblock Learning phrase representations using rnn encoder--decoder for
  statistical machine translation.
\newblock In \emph{Proceedings of the 2014 Conference on Empirical Methods in
  Natural Language Processing (EMNLP)}, pages 1724--1734.

\bibitem[{Covington and McFall(2010)}]{covington2010cutting}
Michael~A Covington and Joe~D McFall. 2010.
\newblock Cutting the gordian knot: The moving-average type--token ratio
  (mattr).
\newblock \emph{Journal of quantitative linguistics}, 17(2):94--100.

\bibitem[{Gao et~al.(2020)Gao, Wang, Herold, Yang, and Ney}]{gao2020towards}
Yingbo Gao, Weiyue Wang, Christian Herold, Zijian Yang, and Hermann Ney. 2020.
\newblock Towards a better understanding of label smoothing in neural machine
  translation.
\newblock In \emph{Proceedings of the 1st Conference of the Asia-Pacific
  Chapter of the Association for Computational Linguistics and the 10th
  International Joint Conference on Natural Language Processing}, pages
  212--223.

\bibitem[{Gu et~al.(2020)Gu, Zhang, Meng, Feng, Xie, Zhou, and
  Yu}]{gu2020token}
Shuhao Gu, Jinchao Zhang, Fandong Meng, Yang Feng, Wanying Xie, Jie Zhou, and
  Dong Yu. 2020.
\newblock Token-level adaptive training for neural machine translation.
\newblock In \emph{Proceedings of the 2020 Conference on Empirical Methods in
  Natural Language Processing (EMNLP)}, pages 1035--1046.

\bibitem[{He et~al.(2019)He, Tu, Wang, Wang, Lyu, and Shi}]{he2019towards}
Shilin He, Zhaopeng Tu, Xing Wang, Longyue Wang, Michael Lyu, and Shuming Shi.
  2019.
\newblock Towards understanding neural machine translation with word
  importance.
\newblock In \emph{Proceedings of the 2019 Conference on Empirical Methods in
  Natural Language Processing and the 9th International Joint Conference on
  Natural Language Processing (EMNLP-IJCNLP)}, pages 952--961.

\bibitem[{Jiang et~al.(2019)Jiang, Ren, Monz, and
  de~Rijke}]{jiang2019improving}
Shaojie Jiang, Pengjie Ren, Christof Monz, and Maarten de~Rijke. 2019.
\newblock Improving neural response diversity with frequency-aware
  cross-entropy loss.
\newblock In \emph{The World Wide Web Conference}, pages 2879--2885.

\bibitem[{Koehn(2004)}]{koehn2004statistical}
Philipp Koehn. 2004.
\newblock Statistical significance tests for machine translation evaluation.
\newblock In \emph{Proceedings of the 2004 conference on empirical methods in
  natural language processing}, pages 388--395.

\bibitem[{Liu et~al.(2021)Liu, Meng, Zhou, Chen, and Xu}]{liu2021faster}
Yijin Liu, Fandong Meng, Jie Zhou, Yufeng Chen, and Jinan Xu. 2021.
\newblock Faster depth-adaptive transformers.
\newblock In \emph{Proceedings of the AAAI Conference on Artificial
  Intelligence}.

\bibitem[{McCarthy and Jarvis(2010)}]{mccarthy2010mtld}
Philip~M McCarthy and Scott Jarvis. 2010.
\newblock Mtld, vocd-d, and hd-d: A validation study of sophisticated
  approaches to lexical diversity assessment.
\newblock \emph{Behavior research methods}, 42(2):381--392.

\bibitem[{Meng and Zhang(2019)}]{meng2019dtmt}
Fandong Meng and Jinchao Zhang. 2019.
\newblock {DTMT:} {A} novel deep transition architecture for neural machine
  translation.
\newblock In \emph{Proceedings of the AAAI Conference on Artificial
  Intelligence}, volume~33, pages 224--231.

\bibitem[{Pereyra et~al.(2017)Pereyra, Tucker, Chorowski, Kaiser, and
  Hinton}]{pereyra2017regularizing}
Gabriel Pereyra, George Tucker, Jan Chorowski, {\L}ukasz Kaiser, and Geoffrey
  Hinton. 2017.
\newblock Regularizing neural networks by penalizing confident output
  distributions.
\newblock \emph{arXiv preprint arXiv:1701.06548}.

\bibitem[{Sennrich et~al.(2016)Sennrich, Haddow, and
  Birch}]{sennrich2016neural}
Rico Sennrich, Barry Haddow, and Alexandra Birch. 2016.
\newblock Neural machine translation of rare words with subword units.
\newblock In \emph{Proceedings of the 54th Annual Meeting of the Association
  for Computational Linguistics (Volume 1: Long Papers)}, pages 1715--1725.

\bibitem[{Sutskever et~al.(2014)Sutskever, Vinyals, and Le}]{NIPS2014seq2seq}
Ilya Sutskever, Oriol Vinyals, and Quoc~V Le. 2014.
\newblock \href
  {https://proceedings.neurips.cc/paper/2014/file/a14ac55a4f27472c5d894ec1c3c743d2-Paper.pdf}
  {Sequence to sequence learning with neural networks}.
\newblock In \emph{Advances in Neural Information Processing Systems},
  volume~27, pages 3104--3112. Curran Associates, Inc.

\bibitem[{Vanmassenhove et~al.(2019)Vanmassenhove, Shterionov, and
  Way}]{vanmassenhove2019lost}
Eva Vanmassenhove, Dimitar Shterionov, and Andy Way. 2019.
\newblock \href {https://www.aclweb.org/anthology/W19-6622} {Lost in
  translation: Loss and decay of linguistic richness in machine translation}.
\newblock In \emph{Proceedings of Machine Translation Summit XVII Volume 1:
  Research Track}, pages 222--232, Dublin, Ireland. European Association for
  Machine Translation.

\bibitem[{Vaswani et~al.(2017)Vaswani, Shazeer, Parmar, Uszkoreit, Jones,
  Gomez, Kaiser, and Polosukhin}]{vaswani2017attention}
Ashish Vaswani, Noam Shazeer, Niki Parmar, Jakob Uszkoreit, Llion Jones,
  Aidan~N. Gomez, L.~Kaiser, and Illia Polosukhin. 2017.
\newblock Attention is all you need.
\newblock In \emph{NIPS}.

\bibitem[{Wan et~al.(2020)Wan, Yang, Wong, Zhou, Chao, Zhang, and
  Chen}]{wan2020self}
Yu~Wan, Baosong Yang, Derek~F Wong, Yikai Zhou, Lidia~S Chao, Haibo Zhang, and
  Boxing Chen. 2020.
\newblock Self-paced learning for neural machine translation.
\newblock In \emph{Proceedings of the 2020 Conference on Empirical Methods in
  Natural Language Processing (EMNLP)}, pages 1074--1080.

\bibitem[{Yan et~al.(2020)Yan, Meng, and Zhou}]{yanetal2020multi}
Jianhao Yan, Fandong Meng, and Jie Zhou. 2020.
\newblock Multi-unit transformers for neural machine translation.
\newblock In \emph{Proceedings of the 2020 Conference on Empirical Methods in
  Natural Language Processing (EMNLP)}, pages 1047--1059, Online.

\bibitem[{Zhang et~al.(2017)Zhang, Ding, Shen, Cheng, Sun, Luan, and
  Liu}]{zhang2017thumt}
Jiacheng Zhang, Yanzhuo Ding, Shiqi Shen, Yong Cheng, Maosong Sun, Huanbo Luan,
  and Yang Liu. 2017.
\newblock Thumt: An open source toolkit for neural machine translation.
\newblock \emph{arXiv preprint arXiv:1706.06415}.

\bibitem[{Zhang et~al.(2019)Zhang, Feng, Meng, You, and
  Liu}]{zhang2019bridging}
Wen Zhang, Yang Feng, Fandong Meng, Di~You, and Qun Liu. 2019.
\newblock Bridging the gap between training and inference for neural machine
  translation.
\newblock In \emph{Proceedings of the 57th Annual Meeting of the Association
  for Computational Linguistics}, pages 4334--4343.

\bibitem[{Zipf(1949)}]{zipf1949human}
George~Kingsley Zipf. 1949.
\newblock Human behavior and the principle of least effort.

\end{thebibliography}

\clearpage

\appendix


\end{document}